%% file: emnlp2020.tex
\documentclass[11pt,a4paper]{article}
\usepackage[hyperref]{emnlp2020}
\usepackage{times}
\usepackage{latexsym}

\usepackage{microtype}

\aclfinalcopy 

\usepackage{graphicx}
\usepackage{booktabs}
\usepackage{caption}
\usepackage{subcaption}
\usepackage{marvosym}
\usepackage{mathtools}
\usepackage{amsmath}
\usepackage{amssymb}
\usepackage{comment}
\usepackage{adjustbox}
\usepackage{lipsum}
\usepackage{xspace}
\newcommand{\ours}{\textsc{TriageSQL}\xspace}

\title{Did You Ask a Good Question? A Cross-Domain Question Intention Classification Benchmark for Text-to-SQL}

\author{
Yusen Zhang$^{1}$\thanks{~~Equal contribution.} \quad
Xiangyu Dong$^{2*}$ \quad
Shuaichen Chang$^{3}$ \\
\bf{
Tao Yu$^{4}$ \quad
Peng Shi$^{5}$ \quad
Rui Zhang$^{6}$ }
\\
 \\
 $^1$ Emory University \quad 
 $^2$ Beihang University \quad 
 $^3$ Ohio State University \\
 $^4$ Yale University \quad
 $^5$ University of Waterloo \quad
 $^6$ Penn State University \\
\small{\tt{yusen.zhang@emory.edu}, \tt{dongxiangyu@buaa.edu.cn}}, \tt{	chang.1692@osu.edu}\\
\small{\tt{tao.yu@yale.edu}, \tt{peng.shi@uwaterloo.ca}, \tt{rmz5227@psu.edu}}
}

\date{}

\begin{document}
\maketitle
\begin{abstract}
Neural models have achieved significant results on the text-to-SQL task, in which most current work assumes all the input questions are legal and generates a SQL query for any input. However, in the real scenario, users can input any text that may not be able to be answered by a SQL query. In this work, we propose \ours, the first cross-domain text-to-SQL question intention classification benchmark that requires models to distinguish four types of unanswerable questions from answerable questions. The baseline RoBERTa model achieves a 60\% F1 score on the test set, demonstrating the need for further improvement on this task. Our dataset is available at \url{https://github.com/chatc/TriageSQL}.
\end{abstract}

\input{introduction}

\input{data}

\input{experiment}

\input{relatedwork}

\input{conclusion}

\bibliographystyle{acl_natbib}
\bibliography{emnlp2020}




\end{document}

%% file: introduction.tex
\section{Introduction}
Text-to-SQL has attracted much attention as a translation task from a natural language question to an executable SQL query. As large-scale datasets are proposed \cite{zhong2017seq,yu2018spider}, recent works have achieved significant results via neural architecture improvements \cite{lyu2020hybrid,Guo2019TowardsCT,rat-sql}. Those works assume all the user inputs are legal, i.e. answerable by a SQL query, which makes models focus on translating input questions to SQL queries. 

\begin{figure}[!th]
    \centering
    \includegraphics[width=0.7\columnwidth]{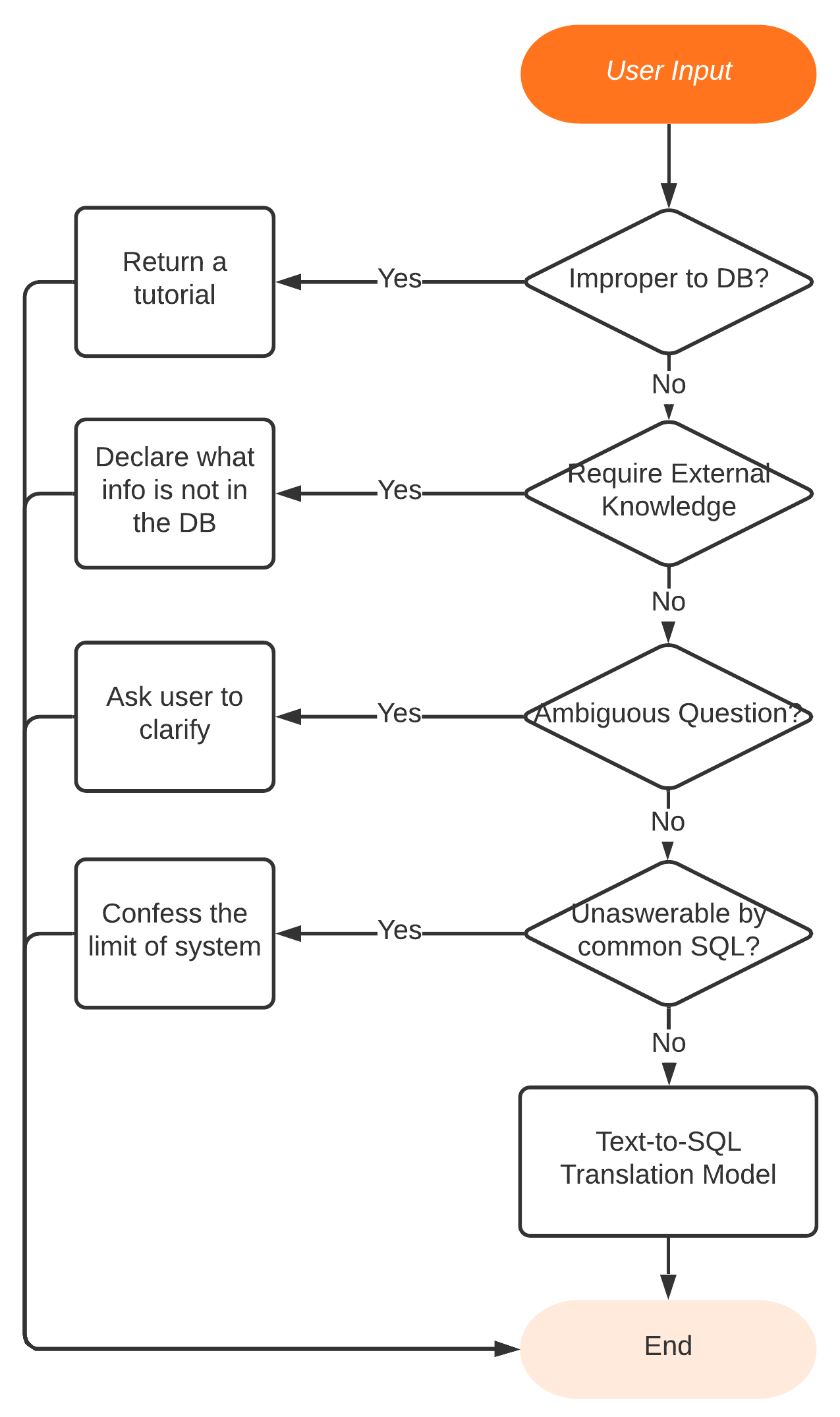}
    \caption{A realistic text-to-SQL system framework where the system distinguishes answerable questions from other possible inputs and then triggers different actions for different types of unanswerable inputs. }
    \label{fig:process}
\end{figure}

Nevertheless, in the real scenario, a text-to-SQL system needs to handle any types of inputs including answerable questions and \emph{unanswerable questions}. In Figure~\ref{fig:process}, we provide different types of user inputs and the corresponding actions of a text-to-SQL system. For answerable questions, the system activates a text-to-SQL translation model to generate SQL queries, while it also needs to rule out all illegal questions with different actions. 

In this work, we propose \ours, a cross-domain text-to-SQL question intention classification benchmark. We formalize four types of unanswerable questions differentiated from answerable questions and construct our benchmark dataset containing 34K databases and 390K questions from 20 existing datasets. We further revised and annotated data to construct a high-quality test set with 500 examples in each type. 
A Transformer-based model achieved a 60\% F1 score when trained and tested on our benchmark for a five-class classification model, which indicates the challenge of this task.

\begin{table*}[ht!]
\centering
\resizebox{\textwidth}{!}{%
    \begin{tabular}{l|l|l|l}
    \toprule
     
    Data Type &  Sample Question & Table Information and Descriptions & Source\\
    \midrule
    Improper & But they say they don't have any human side effects, hum! & The schema is not related to the question. & Alexa\\
    ExtKnow & Which nation is Danilo Di Luca from? & Removing column ``nation'' from original tables. & WikiSQL \\ 
    Ambiguous & Who is the coach of UCLA? & Schema contains both ``season coach'' and ``interim head coach''. & WikiSQL \\
    Non-SQL & Who is the first away team on the chart? & Requiring the row positions which is not listed in tables. & WTQ\\
    Answerable & How many drivers are there? & A schema with Driver, School, School Bus. & Spider \\
    \bottomrule
    \end{tabular}%
}
\caption{Samples of five types in \ours. The third column shows the table information revealing the reason for being classified as its data type. The last column shows the origins of the questions before they are modified.}
\label{table:sample}
\end{table*}

We summarize our contributions as below:

1) To the best of our knowledge, this work is the first to focus on the unanswerable questions in the text-to-SQL task. We construct a five-class question classification task with 390K pairs of questions and databases.

2) Besides the training set we collect, a high-quality test set is also presented after manual annotation. Future works can follow our training set and improve models or construct their own training data and test on our test set.

3) A strong baseline model reveals the challenge of this task, and the analysis of results demonstrates how each type of questions confuses the model.

%% file: data.tex
\section{Task Description}

This task aims to distinguish answerable questions from unanswerable inputs. Given a database schema and a natural language text, a model is to determine the type of the text based on the relation of the database schema and the text. We formalize the following five types of input, where different types of inputs correspond to different actions of a text-to-SQL system (Figure~\ref{fig:process}). Table~\ref{table:sample} displays some examples of each type.

\paragraph{Improper to DB} (\texttt{Improper})
Some questions such as small talk or asking-opinion questions are not proper to any databases. Apparently, users who asked those questions are not familiar with the function of databases, so the best strategy of the system is to return a tutorial instead of trying to answer those questions.

\paragraph{Require external knowledge} (\texttt{ExtKnow})
Users are likely to ask unanswerable questions because the questions require extra information which is not in the database. This is common especially when users are not familiar with the database schema or its meaning. The following action of a system is to detect what information is required by the user but not included in the database.

\paragraph{Ambiguous}  (\texttt{Ambiguous})
The system may not answer the question correctly due to the ambiguity of user questions. These ambiguous questions can usually be paraphrased into more than one SQL parses in our settings, because they contain some words not explicitly pointing to either 1) column/table names (ambiguity in schema elements), such as finding the ``name'' of a person for a schema containing ``last name'' as well as ``first name''; or 2) conditions of operators (ambiguity in values), such as finding the ``good'' movie while ``good'' is an ambiguous measurement for selecting a correct value. The system should indicate the ambiguous part and ask the user to clarify.

\paragraph{Unanswerable by common SQL grammar} (\texttt{Non-SQL})
This type contains the questions and corresponding queries that are not executable by text-to-SQL systems with common SQL grammar, i.e. the grammar only containing operators in the current mainstream text-to-SQL systems \cite{zhong2017seq,yu2018spider}, due to the limitation of current text-to-SQL datasets and models.

First, this type contains some questions that are totally unanswerable by any SQL grammar, such as declarative statements and multi-modality questions. Furthermore, we also add some questions beyond the reach of current text-to-SQL systems because they are answerable only by uncommon grammar. For example, as shown in Table~\ref{table:sample}, users may query information about the database table itself, such as column/table names and the order of records, or about the return values of SQL queries. Since such questions are reasonable questions but out of the capability of text-to-SQL models, the system needs to know its limit.

\paragraph{Answerable} (\texttt{Answerable})
This type refers to regular questions that a text-to-SQL translation model can handle. 


\begin{table*}[t]
\centering
\resizebox{\textwidth}{!}{%
    \begin{tabular}{lccccccc}
    \toprule
     & \shortstack{Spider \\ SparC, CoSQL} & WikiSQL & \shortstack{Restaurants,Scholar,Yelp\\IMDB,Geo,Academic} & \shortstack{TabFact,ToTTo\\ LogicNLG} & HybridQA & WTQ & \shortstack{Alexa, NQ, MARCO \\ WikiQA, CoQA, QuAC} \\
    \midrule
    Improper   &  -    &    -  &   -    &  -    &   -   &   -    & 48689 \\
    ExtKnow    & 14120 & 80654 & 1315  &    -   & 39819 &    -   &   -   \\
    Ambiguous  & 962  &   -   & 760   &    -   &   -   &   -    &  -    \\
    Non-SQL    &  -    &   -   &   -    &   169018    &   -   & 14152  &   -   \\
    Answerable & 48132 & 161308 & 4706  &  -     &  -    &     -  &  -    \\
    \bottomrule
    \end{tabular}
}
\caption{The source of each question type in \ours.}
\label{table:data-source}
\end{table*}


\section{Dataset Construction}
We construct our benchmark from 21 text-to-SQL, question answering, and table-to-text datasets. Table~\ref{table:data-source} summarizes the dataset source of each question type, and we briefly describe our construction procedure below.

\subsection{Data Sources and Construction Methods}
\paragraph{Spider, SparC, CoSQL} are cross-domain text-to-SQL datasets containing 162 different databases and corresponding questions \cite{yu2018spider,yu2019sparc,yu2019cosql}.
Spider is single-turn, while the other two are multi-turn.
To collect \texttt{Answerable} questions, we collect questions from Spider and the first-turn questions from SparC and CoSQL since we focus on context-independent questions in this work.
We also collect the \texttt{Ambiguous} questions annotated in CoSQL. In addition, the schemas of these dataset are paired with the chit-chat quesitons, e.g. Alexa, to form \texttt{Improper} questions.
Furthermore, we construct \texttt{ExtKnow} questions by ablating the database schema. 

To build \texttt{ExtKnow} questions, we randomly remove 1-3 columns that are mentioned in SQL from the database schema, so that the database does not contain sufficient information to answer the question.
To prevent models from recognizing \texttt{ExtKnow} questions by simply detecting incomplete database schema, we also add some answerable questions by removing 1-3 unmentioned columns from the schema.
In Section~\ref{sec:discussion}, we will show that these examples make it particularly challenging for the model to distinguish \texttt{ExtKnow} and \texttt{Answerable} questions.


\paragraph{WikiSQL} is a cross-domain text-to-SQL dataset over Wikipedia tables \cite{zhong2017seq}. 
WikiSQL is also used to construct \texttt{Answerable} and \texttt{ExtKnow} questions in a similar way to Spider/SparC/CoSQL. 

In addition, WikiSQL contributes to the construction of  \texttt{Ambiguous} questions as well. To achieve this, we firstly apply two rules to generate the candidates of ambiguous questions. The first rule is to use the overlapped words between two columns in a schema to replace a column name in question, e.g. ``away team'' in question would be changed to ``team'' since both ``home team'' as well as ``away team'' are in the schema. The second rule is simply to use some ambiguous concepts to replace the numbers or strings in the questions, such as changing ``15'' to ``around 15''.  Then, secondly, we further label about 200 samples from the generated candidates manually to make sure the correctness.

\paragraph{Restaurants, Scholar, Yelp, IMDB} are four single-database text-to-SQL datasets where some questions are ambiguous during their annotation process \cite{tang2000automated,iyer2017learning,yaghmazadeh2017type,acl18sql}. 
We first manually annotate the ambiguous questions in these four datasets to construct \texttt{Ambiguous} questions. Then, the remaining questions are used to construct \texttt{Answerable} and \texttt{ExtKnow} questions.

\paragraph{TabFact, ToTTo, LogicNLG} are table-to-text datasets where the text is the description or statement of a part of the database information instead of questions to them \cite{2019TabFactA,parikh2020totto,Chen2020LogicalNL}. 
Therefore, we extract the text-schema pairs in these three datasets as \texttt{Non-SQL} type because they are related to the database but not answerable by SQL. 

\paragraph{HybridQA}  
is a question answering dataset requiring the combination of a Wikipedia table and a paragraph to answer a question \cite{chen2020hybridqa}.
We generate \texttt{ExtKnow} questions naturally by removing the needed passages for inferring correct answers. In this way, these questions become unanswerable due to the lack of information out of databases. 

\paragraph{WikiTableQuestions (WTQ)} is a question answering dataset over Wikipedia tables annotated with corresponding answers instead of SQL \cite{pasupat2015compositional}. Some of its questions cannot be answered by SQL or require uncommon SQL grammar which is never covered by any existing text-to-SQL datasets.
Therefore, WTQ is mainly used for constructing \texttt{Non-SQL} questions. 

Similar to ambiguous question generation, we first use two rules to detect such questions. The first one is to extract the questions with string-type answers not appearing in the content of databases, since most common grammar is not able to return the string type values other than the content of database. The second one is to extract the questions related to the information about the DB itself using keyword matching, such as ``chart'' and ``row''.
    
\paragraph{Alexa, Natural Questions (NQ), MARCO, WikiQA, CoQA, QuAC} contains chit-chat questions (or statements) of passages, concepts, documents, and web searches instead of relational databases \cite{Gopalakrishnan2019,kwiatkowski2019natural,nguyen2016ms,yang2015wikiqa,reddy2019coqa,choi2018quac}.
Therefore, all of these questions are not answerable by SQL, so we use them to construct \texttt{Improper} questions.

To make sure the consistency with other type of questions, we sample 1/10 of questions from these dataset. For each question, we randomly choose a schema from datasets containing DBs, e.g. Spider, as the corresponding schema of the question.

\subsection{Train, Dev, Test split}
\label{sec:split}
To make \ours a cross-domain dataset, we first merge the databases from different sources with the same schema and their corresponding questions. Second, we randomly select 80\% of databases for training and development set and leave the rest 20\% as test set candidates (Table.\ref{table:data-count}). We split questions based on schema, which means there is no overlap of schema in train, development, and test set. Finally, two students with SQL expertise manually pick 500 high-quality examples from test set candidates for each  type (including manually labeled extra 200 ambiguous questions from WikiSQL), 2500 examples in total and another 2 students confirm the examples in test set. In this way, this test set is sampled and then revised by humans to make it high-quality and representative.
\begin{table}[t]
\centering
    \begin{tabular}{lccc}
    \toprule
     & Train & Dev & Test \\
    \midrule
    Improper   &  31160 & 7790 & 500 \\
    ExtKnow    &  90328 & 19546 & 500 \\ 
    Ambiguous  &  956 & 189 & 500 \\ 
    Non-SQL    &  122531 & 25992 & 500\\ 
    Answerable &  139884 & 32892 & 500\\ 
    \midrule
    Schema & 87601 & 22566& 1470\\
    \bottomrule
    \end{tabular}
\caption{Data Splits and Statistics in \ours. First 5 rows show the number of questions while the last row shows the number of schema. Test indicates manually labeled test data set.} 
\label{table:data-count}
\end{table}



%% file: experiment.tex
\begin{table}[ht!]
  \centering
  \begin{tabular}{lccc}
    \toprule
     & Precision & Recall & Macro F1    \\ 
        \midrule
    Improper   & 0.79       & 0.88   & 0.83            \\
    ExtKnow    & 0.69      & 0.45   & 0.54           \\
    Ambiguous  & 0.81      & 0.09    & 0.17         \\ 
    Non-SQL    & 0.91      & 0.92   & 0.92       \\
    Answerable & 0.39       & 0.82   & 0.53          \\
    \midrule
    Average    & 0.72      & 0.63   & 0.60         \\ 
    \bottomrule
  \end{tabular}
  \caption{RoBERTa model on the \ours test set.}
  \label{tab:result}
\end{table}
\begin{figure}[ht!]
    \centering
    \includegraphics[width=\linewidth]{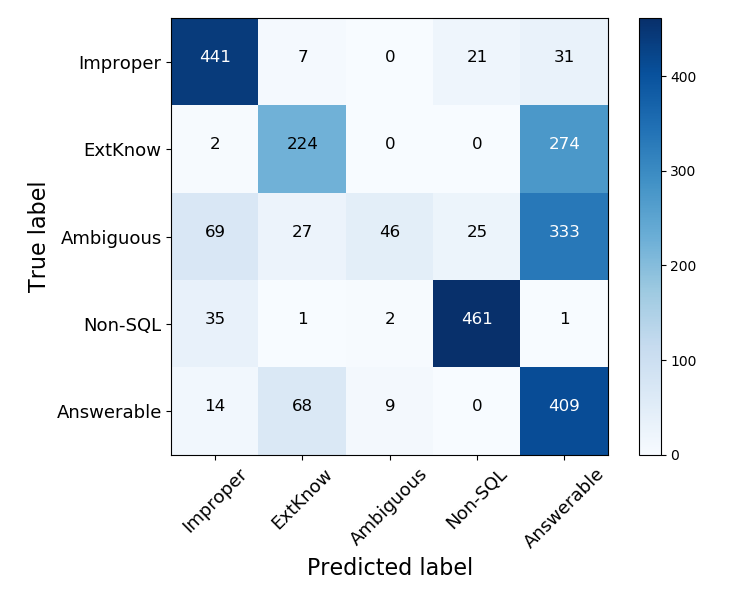}
    \caption{Confusion matrix of the RoBERTa model on test set. The horizontal axis indicates the predicted labels while the vertical axis indicates the true labels. The number in row $i$, column $j$ represents the number of samples in type $i$ that are classified as type $j$.}
    \label{fig:confusion}
\end{figure}
\section{Experiment and Result}

\subsection{Model}
We use RoBERTa \cite{liu2019roberta} as the baseline model to test the difficulty of our task. RoBERTa is a strong pre-trained language model based on Transformers \cite{vaswani2017attention,devlin2018bert} which has been widely used in encoding text  and DB schema \cite{hwang2019comprehensive}. Given a pair of question and schema, we use a special token to separate question and each column in the schema and fine-tune a pre-trained RoBERTa-base model on a sampled training set with at most 10k samples in each type. 

\subsection{Result}
Table~\ref{tab:result} shows the result of the RoBERTa model on the proposed test set, achieving a 60\% F1 score on average.
Some question types can be classified with high F1 scores, such as \texttt{Improper} questions and \texttt{Non-SQL} questions unanswerable by common SQL grammar.
However, it only obtains 17\% F1 score on \texttt{Ambiguous} questions. The lower scores could be the result of inherent difficulty of ambiguous questions. 

To further analyze the distribution of prediction, Figure~\ref{fig:confusion} displays the confusion matrix of the results. As shown in the figure, the model is likely confused with \texttt{ExtKnow} and \texttt{Answerable} questions, demonstrating that it is challenging to figure out the missing column in the database and classify them as unanswerable questions. In addition to this, ambiguous questions are relatively hard to be classified as the correct type, demonstrating the difficulties of our task.

\subsection{Discussion}
\label{sec:discussion}
\begin{table}[ht!]
  \centering
  \begin{tabular}{lccc}
    \toprule
          & Precision & Recall & F1 \\
    \midrule
    \multicolumn{4}{l}{w/ non-mentioned} \\
    Answerable   & 0.55       & 0.73   & 0.63     \\
    ExtKnow    & 0.63      & 0.41   & 0.49          \\
    \midrule
    \multicolumn{4}{l}{w/o non-mentioned} \\
    Answerable  & 0.69      & 0.88    & 0.78         \\ 
    ExtKnow    & 0.87      & 0.64   & 0.74 \\ 
    \bottomrule
  \end{tabular}
  \caption{Scores of \texttt{ExtKnow} and \texttt{Answerable} questions on the development set. w/ non-mentioned is the dataset containing the answerable questions with non-mentioned column removed, while w/o non-mentioned does not contains. }
  \label{tab:ablation}
\end{table}

To further explore the factors of confusion between \texttt{ExtKnow} and \texttt{Answerable} questions, we hypothesize that the \texttt{Answerable} questions constructed by removing one to three non-mentioned columns make it challenging for the model to distinguish the two types.
To test this, we modify the dataset by deleting the samples in \texttt{Answerable} questions which are constructed by removing one to three non-mentioned columns, and then evaluate the model on both the original and this modified dataset. For each dataset, We randomly sample a development set (10k samples for each type).
As shown in Table~\ref{tab:ablation}, the model evaluated on the original dataset obtains lower scores although it uses more training data. This result demonstrates that non-mentioned columns are able to confuse the model. Adding this type of data into the dataset would largely increase the difficulty of the task.

%% file: relatedwork.tex
\section{Related Work}

While Text-to-SQL and natural language interface to database systems have been studied for decades \cite{warren1982efficient,miller1996fully,popescu2003towards,li2014constructing,zhong2017seq,acl18sql,yu2018spider}, most of the prior work assume the user question is answerable by SQL queries and can be handled by the system.
The only text-to-SQL work to study out-of-scope user questions is CoSQL \cite{yu2019cosql} where they perform dialog act classification to identify ambiguous and answerable questions in a conversational setting.
Furthermore, question intent classification and out-of-scope detection has been studies for task-oriented dialog systems \cite{braun2017evaluating,coucke2018snips,larson2019evaluation,zheng2020out,yilmaz2020kloos,feng2020none,casanueva2020efficient,larson2020iterative}, community question answering \cite{chen2012understanding}, and machine comprehension \cite{rajpurkar2018know}.
By contrast, our paper focuses on the cross-domain text-to-SQL task and proposes the first benchmark to provide a systematic taxonomy and comprehensive analysis of different question types for text-to-SQL systems.

%% file: conclusion.tex
\section{Conclusion}
We proposed \ours, a cross-domain benchmark for text-to-SQL intention classification. In addition to the answerable questions, it contains four types of unanswerable questions to help text-to-SQL systems deal with different inputs. We use a baseline RoBERTa model to measure the difficulty of human refined test set. The result demonstrates a significant space for improvement. The future study includes collecting naturally occurring examples of questions that fit our criteria \cite{de2020towards}, reducing the training and test set gap, and creating better models. Also, another direction is to generate suggestions by the system based on the classification results, e.g. provide users the missed columns or the required knowledge in \texttt{ExtKnow} questions.

\section{Acknowledgments}
We thank Chang Shu for his help with dataset construction. We also thank the anonymous reviewers for their valuable feedback.

%% file: emnlp2020.bbl
\begin{thebibliography}{41}
\expandafter\ifx\csname natexlab\endcsname\relax\def\natexlab#1{#1}\fi

\bibitem[{amd Anthony~Zheng et~al.(2020)amd Anthony~Zheng, Mahendran, Tekriwal,
  Cheung, Guldan, Leach, and Kummerfeld}]{larson2020iterative}
Stefan~Larson amd Anthony~Zheng, Anish Mahendran, Rishi Tekriwal, Adrian
  Cheung, Eric Guldan, Kevin Leach, and Jonathan~K. Kummerfeld. 2020.
\newblock Iterative feature mining for constraint-based data collection to
  increase data diversity and model robustness.
\newblock In \emph{Proceedings of the 2020 Conference on Empirical Methods in
  Natural Language Processing}.

\bibitem[{Braun et~al.(2017)Braun, Mendez, Matthes, and
  Langen}]{braun2017evaluating}
Daniel Braun, Adrian~Hernandez Mendez, Florian Matthes, and Manfred Langen.
  2017.
\newblock \href {https://www.aclweb.org/anthology/W17-5522} {Evaluating natural
  language understanding services for conversational question answering
  systems}.
\newblock In \emph{Proceedings of the 18th Annual SIGdial Meeting on Discourse
  and Dialogue}, pages 174--185.

\bibitem[{Casanueva et~al.(2020)Casanueva, Tem{\v{c}}inas, Gerz, Henderson, and
  Vuli{\'c}}]{casanueva2020efficient}
I{\~n}igo Casanueva, Tadas Tem{\v{c}}inas, Daniela Gerz, Matthew Henderson, and
  Ivan Vuli{\'c}. 2020.
\newblock \href {https://www.aclweb.org/anthology/2020.nlp4convai-1.5}
  {Efficient intent detection with dual sentence encoders}.
\newblock In \emph{Proceedings of the 18th Annual SIGdial Meeting on Discourse
  and Dialogue}, pages 38--45.

\bibitem[{Chen et~al.(2012)Chen, Zhang, and Mark}]{chen2012understanding}
Long Chen, Dell Zhang, and Levene Mark. 2012.
\newblock \href {https://dl.acm.org/doi/abs/10.1145/2187980.2188206}
  {Understanding user intent in community question answering}.
\newblock In \emph{Proceedings of the 21st international conference on world
  wide web}, pages 823--828.

\bibitem[{Chen et~al.(2020{\natexlab{a}})Chen, Chen, Su, Chen, and
  Wang}]{Chen2020LogicalNL}
Wenhu Chen, Jianshu Chen, Yu~Su, Zhiyu Chen, and William~Yang Wang.
  2020{\natexlab{a}}.
\newblock \href {https://www.aclweb.org/anthology/2020.acl-main.708} {Logical
  natural language generation from open-domain tables}.
\newblock In \emph{Proceedings of the 58th Annual Meeting of the Association
  for Computational Linguistics}, pages 7929--7942.

\bibitem[{Chen et~al.(2020{\natexlab{b}})Chen, Wang, Chen, Zhang, Wang, Li,
  Zhou, and Wang}]{2019TabFactA}
Wenhu Chen, Hongmin Wang, Jianshu Chen, Yunkai Zhang, Hong Wang, Shiyang Li,
  Xiyou Zhou, and William~Yang Wang. 2020{\natexlab{b}}.
\newblock \href {https://openreview.net/forum?id=rkeJRhNYDH} {{TabFact} : A
  large-scale dataset for table-based fact verification}.
\newblock In \emph{International Conference on Learning Representations
  (ICLR)}.

\bibitem[{Chen et~al.(2020{\natexlab{c}})Chen, Zha, Chen, Xiong, Wang, and
  Wang}]{chen2020hybridqa}
Wenhu Chen, Hanwen Zha, Zhiyu Chen, Wenhan Xiong, Hong Wang, and William Wang.
  2020{\natexlab{c}}.
\newblock \href {https://arxiv.org/abs/2004.07347} {Hybridqa: A dataset of
  multi-hop question answering over tabular and textual data}.
\newblock \emph{arXiv preprint arXiv:2004.07347}.

\bibitem[{Choi et~al.(2018)Choi, He, Iyyer, Yatskar, Yih, Choi, Liang, and
  Zettlemoyer}]{choi2018quac}
Eunsol Choi, He~He, Mohit Iyyer, Mark Yatskar, Wen-tau Yih, Yejin Choi, Percy
  Liang, and Luke Zettlemoyer. 2018.
\newblock \href {https://www.aclweb.org/anthology/D18-1241} {{Q}u{AC}: Question
  answering in context}.
\newblock In \emph{Proceedings of the 2018 Conference on Empirical Methods in
  Natural Language Processing}, pages 2174--2184.

\bibitem[{Coucke et~al.(2018)Coucke, Saade, Ball, Bluche, Caulier, Leroy,
  Doumouro, Gisselbrecht, Caltagirone, Lavril et~al.}]{coucke2018snips}
Alice Coucke, Alaa Saade, Adrien Ball, Th{\'e}odore Bluche, Alexandre Caulier,
  David Leroy, Cl{\'e}ment Doumouro, Thibault Gisselbrecht, Francesco
  Caltagirone, Thibaut Lavril, et~al. 2018.
\newblock \href {https://arxiv.org/abs/1805.10190} {Snips voice platform: an
  embedded spoken language understanding system for private-by-design voice
  interfaces}.
\newblock \emph{arXiv preprint arXiv:1805.10190}.

\bibitem[{Devlin et~al.(2019)Devlin, Chang, Lee, and
  Toutanova}]{devlin2018bert}
Jacob Devlin, Ming-Wei Chang, Kenton Lee, and Kristina Toutanova. 2019.
\newblock \href {https://www.aclweb.org/anthology/N19-1423} {{BERT}:
  Pre-training of deep bidirectional transformers for language understanding}.
\newblock In \emph{Proceedings of the 2019 Conference of the North {A}merican
  Chapter of the Association for Computational Linguistics}, pages 4171--4186.

\bibitem[{Feng et~al.(2020)Feng, Mehri, Eskenazi, and Zhao}]{feng2020none}
Yulan Feng, Shikib Mehri, Maxine Eskenazi, and Tiancheng Zhao. 2020.
\newblock \href {https://www.aclweb.org/anthology/2020.acl-main.182} {`` {None
  of the Above} '': Measure uncertainty in dialog response retrieval}.
\newblock In \emph{Proceedings of the 58th Annual Meeting of the Association
  for Computational Linguistics}, pages 2013--2020.

\bibitem[{Finegan-Dollak et~al.(2018)Finegan-Dollak, Kummerfeld, Zhang,
  Ramanathan, Sadasivam, Zhang, and Radev}]{acl18sql}
Catherine Finegan-Dollak, Jonathan~K. Kummerfeld, Li~Zhang, Karthik Ramanathan,
  Sesh Sadasivam, Rui Zhang, and Dragomir Radev. 2018.
\newblock \href {http://www.aclweb.org/anthology/P18-1033.pdf} {Improving
  text-to-sql evaluation methodology}.
\newblock In \emph{Proceedings of the 56th Annual Meeting of the Association
  for Computational Linguistics}, pages 351--360.

\bibitem[{Gopalakrishnan et~al.(2019)Gopalakrishnan, Hedayatnia, Chen,
  Gottardi, Kwatra, Venkatesh, Gabriel, and Hakkani-Tür}]{Gopalakrishnan2019}
Karthik Gopalakrishnan, Behnam Hedayatnia, Qinlang Chen, Anna Gottardi, Sanjeev
  Kwatra, Anu Venkatesh, Raefer Gabriel, and Dilek Hakkani-Tür. 2019.
\newblock \href {http://dx.doi.org/10.21437/Interspeech.2019-3079}
  {{Topical-Chat: Towards Knowledge-Grounded Open-Domain Conversations}}.
\newblock In \emph{Proceedings of Interspeech 2019}, pages 1891--1895.

\bibitem[{Guo et~al.(2019)Guo, Zhan, Gao, Xiao, Lou, Liu, and
  Zhang}]{Guo2019TowardsCT}
Jiaqi Guo, Zecheng Zhan, Yan Gao, Yan Xiao, Jian-Guang Lou, Ting Liu, and
  Dongmei Zhang. 2019.
\newblock \href {https://www.aclweb.org/anthology/P19-1444} {Towards complex
  text-to-{SQL} in cross-domain database with intermediate representation}.
\newblock In \emph{Proceedings of the 57th Annual Meeting of the Association
  for Computational Linguistics}, pages 4524--4535.

\bibitem[{Hwang et~al.(2019)Hwang, Yim, Park, and Seo}]{hwang2019comprehensive}
Wonseok Hwang, Jinyeong Yim, Seunghyun Park, and Minjoon Seo. 2019.
\newblock \href {https://arxiv.org/abs/1902.01069} {A comprehensive exploration
  on wikisql with table-aware word contextualization}.
\newblock \emph{arXiv preprint arXiv:1902.01069}.

\bibitem[{Iyer et~al.(2017)Iyer, Konstas, Cheung, Krishnamurthy, and
  Zettlemoyer}]{iyer2017learning}
Srinivasan Iyer, Ioannis Konstas, Alvin Cheung, Jayant Krishnamurthy, and Luke
  Zettlemoyer. 2017.
\newblock \href {https://www.aclweb.org/anthology/P17-1089} {Learning a neural
  semantic parser from user feedback}.
\newblock In \emph{Proceedings of the 55th Annual Meeting of the Association
  for Computational Linguistics}, pages 963--973.

\bibitem[{Kwiatkowski et~al.(2019)Kwiatkowski, Palomaki, Redfield, Collins,
  Parikh, Alberti, Epstein, Polosukhin, Devlin, Lee
  et~al.}]{kwiatkowski2019natural}
Tom Kwiatkowski, Jennimaria Palomaki, Olivia Redfield, Michael Collins, Ankur
  Parikh, Chris Alberti, Danielle Epstein, Illia Polosukhin, Jacob Devlin,
  Kenton Lee, et~al. 2019.
\newblock \href
  {https://www.mitpressjournals.org/doi/full/10.1162/tacl_a_00276} {Natural
  questions: a benchmark for question answering research}.
\newblock \emph{Transactions of the Association for Computational Linguistics},
  pages 453--466.

\bibitem[{Larson et~al.(2019)Larson, Mahendran, Peper, Clarke, Lee, Hill,
  Kummerfeld, Leach, Laurenzano, Tang, and Mars}]{larson2019evaluation}
Stefan Larson, Anish Mahendran, Joseph~J. Peper, Christopher Clarke, Andrew
  Lee, Parker Hill, Jonathan~K. Kummerfeld, Kevin Leach, Michael~A. Laurenzano,
  Lingjia Tang, and Jason Mars. 2019.
\newblock \href {https://www.aclweb.org/anthology/D19-1131} {An evaluation
  dataset for intent classification and out-of-scope prediction}.
\newblock In \emph{Proceedings of the 2019 Conference on Empirical Methods in
  Natural Language Processing}, pages 1311--1316.

\bibitem[{Li and Jagadish(2014)}]{li2014constructing}
Fei Li and HV~Jagadish. 2014.
\newblock \href {https://dl.acm.org/doi/abs/10.14778/2735461.2735468}
  {Constructing an interactive natural language interface for relational
  databases}.
\newblock \emph{Proceedings of the VLDB Endowment}, pages 73--84.

\bibitem[{Liu et~al.(2019)Liu, Ott, Goyal, Du, Joshi, Chen, Levy, Lewis,
  Zettlemoyer, and Stoyanov}]{liu2019roberta}
Yinhan Liu, Myle Ott, Naman Goyal, Jingfei Du, Mandar Joshi, Danqi Chen, Omer
  Levy, Mike Lewis, Luke Zettlemoyer, and Veselin Stoyanov. 2019.
\newblock \href {https://arxiv.org/abs/1907.11692} {{RoBERTa}: A robustly
  optimized bert pretraining approach}.
\newblock \emph{arXiv preprint arXiv:1907.11692}.

\bibitem[{Lyu et~al.(2020)Lyu, Chakrabarti, Hathi, Kundu, Zhang, and
  Chen}]{lyu2020hybrid}
Qin Lyu, Kaushik Chakrabarti, Shobhit Hathi, Souvik Kundu, Jianwen Zhang, and
  Zheng Chen. 2020.
\newblock \href
  {https://www.microsoft.com/en-us/research/publication/hybrid-ranking-network-for-text-to-sql/}
  {Hybrid ranking network for text-to-sql}.
\newblock Technical Report MSR-TR-2020-7, Microsoft Dynamics 365 AI.

\bibitem[{Miller et~al.(1996)Miller, Stallard, Bobrow, and
  Schwartz}]{miller1996fully}
Scott Miller, David Stallard, Robert Bobrow, and Richard Schwartz. 1996.
\newblock \href {https://www.aclweb.org/anthology/P96-1008} {A fully
  statistical approach to natural language interfaces}.
\newblock In \emph{34th Annual Meeting of the Association for Computational
  Linguistics}, pages 55--61.

\bibitem[{Nguyen et~al.(2016)Nguyen, Rosenberg, Song, Gao, Tiwary, Majumder,
  and Deng}]{nguyen2016ms}
Tri Nguyen, Mir Rosenberg, Xia Song, Jianfeng Gao, Saurabh Tiwary, Rangan
  Majumder, and Li~Deng. 2016.
\newblock \href
  {https://www.microsoft.com/en-us/research/publication/ms-marco-human-generated-machine-reading-comprehension-dataset/}
  {Ms marco: A human generated machine reading comprehension dataset}.
\newblock \emph{arXiv preprint arXiv:1611.09268}.

\bibitem[{Parikh et~al.(2020)Parikh, Wang, Gehrmann, Faruqui, Dhingra, Yang,
  and Das}]{parikh2020totto}
Ankur~P Parikh, Xuezhi Wang, Sebastian Gehrmann, Manaal Faruqui, Bhuwan
  Dhingra, Diyi Yang, and Dipanjan Das. 2020.
\newblock \href {https://arxiv.org/abs/2004.14373} {Totto: A controlled
  table-to-text generation dataset}.
\newblock In \emph{Proceedings of the 2020 Conference on EmpiricalMethods in
  Natural Language Processing}.

\bibitem[{Pasupat and Liang(2015)}]{pasupat2015compositional}
Panupong Pasupat and Percy Liang. 2015.
\newblock \href {https://www.aclweb.org/anthology/P15-1142} {Compositional
  semantic parsing on semi-structured tables}.
\newblock In \emph{Proceedings of the 53rd Annual Meeting of the Association
  for Computational Linguistics}, pages 1470--1480.

\bibitem[{Popescu et~al.(2003)Popescu, Etzioni, and Kautz}]{popescu2003towards}
Ana-Maria Popescu, Oren Etzioni, and Henry Kautz. 2003.
\newblock \href {https://dl.acm.org/doi/abs/10.1145/604045.604070} {Towards a
  theory of natural language interfaces to databases}.
\newblock In \emph{Proceedings of the 8th international conference on
  Intelligent user interfaces}, pages 149--157.

\bibitem[{Rajpurkar et~al.(2018)Rajpurkar, Jia, and Liang}]{rajpurkar2018know}
Pranav Rajpurkar, Robin Jia, and Percy Liang. 2018.
\newblock \href {https://www.aclweb.org/anthology/P18-2124} {Know what you
  don{'}t know: Unanswerable questions for {SQ}u{AD}}.
\newblock In \emph{Proceedings of the 56th Annual Meeting of the Association
  for Computational Linguistics}, pages 784--789.

\bibitem[{Reddy et~al.(2019)Reddy, Chen, and Manning}]{reddy2019coqa}
Siva Reddy, Danqi Chen, and Christopher~D Manning. 2019.
\newblock \href
  {https://www.mitpressjournals.org/doi/full/10.1162/tacl_a_00266} {Coqa: A
  conversational question answering challenge}.
\newblock \emph{Transactions of the Association for Computational Linguistics},
  pages 249--266.

\bibitem[{Tang and Mooney(2000)}]{tang2000automated}
Lappoon~R. Tang and Raymond~J. Mooney. 2000.
\newblock \href {https://www.aclweb.org/anthology/W00-1317} {Automated
  construction of database interfaces: Intergrating statistical and relational
  learning for semantic parsing}.
\newblock In \emph{2000 Joint {SIGDAT} Conference on Empirical Methods in
  Natural Language Processing and Very Large Corpora}, pages 133--141.

\bibitem[{Vaswani et~al.(2017)Vaswani, Shazeer, Parmar, Uszkoreit, Jones,
  Gomez, Kaiser, and Polosukhin}]{vaswani2017attention}
Ashish Vaswani, Noam Shazeer, Niki Parmar, Jakob Uszkoreit, Llion Jones,
  Aidan~N Gomez, {\L}ukasz Kaiser, and Illia Polosukhin. 2017.
\newblock \href {https://arxiv.org/abs/1706.03762} {Attention is all you need}.
\newblock In \emph{Advances in neural information processing systems}, pages
  5998--6008.

\bibitem[{de~Vries et~al.(2020)de~Vries, Bahdanau, and Manning}]{de2020towards}
Harm de~Vries, Dzmitry Bahdanau, and Christopher Manning. 2020.
\newblock \href {https://arxiv.org/abs/2007.14435} {Towards ecologically valid
  research on language user interfaces}.
\newblock \emph{arXiv preprint arXiv:2007.14435}.

\bibitem[{Wang et~al.(2020)Wang, Shin, Liu, Polozov, and Richardson}]{rat-sql}
Bailin Wang, Richard Shin, Xiaodong Liu, Oleksandr Polozov, and Matthew
  Richardson. 2020.
\newblock \href {https://www.aclweb.org/anthology/2020.acl-main.677}
  {{RAT-SQL}: Relation-aware schema encoding and linking for text-to-{SQL}
  parsers}.
\newblock In \emph{Proceedings of the 58th Annual Meeting of the Association
  for Computational Linguistics}, pages 7567--7578.

\bibitem[{Warren and Pereira(1982)}]{warren1982efficient}
David~HD Warren and Fernando~CN Pereira. 1982.
\newblock \href {https://dl.acm.org/doi/abs/10.5555/972942.972944} {An
  efficient easily adaptable system for interpreting natural language queries}.
\newblock \emph{Computational Linguistics}, pages 110--122.

\bibitem[{Yaghmazadeh et~al.(2017)Yaghmazadeh, Wang, Dillig, and
  Dillig}]{yaghmazadeh2017type}
Navid Yaghmazadeh, Yuepeng Wang, Isil Dillig, and Thomas Dillig. 2017.
\newblock \href {https://arxiv.org/abs/1702.01168} {Type-and content-driven
  synthesis of sql queries from natural language}.
\newblock \emph{arXiv preprint arXiv:1702.01168}.

\bibitem[{Yang et~al.(2015)Yang, Yih, and Meek}]{yang2015wikiqa}
Yi~Yang, Wen-tau Yih, and Christopher Meek. 2015.
\newblock \href {https://www.aclweb.org/anthology/D15-1237} {{W}iki{QA}: A
  challenge dataset for open-domain question answering}.
\newblock In \emph{Proceedings of the 2015 Conference on Empirical Methods in
  Natural Language Processing}, pages 2013--2018.

\bibitem[{Yilmaz and Toraman(2020)}]{yilmaz2020kloos}
Eyup~Halit Yilmaz and Cagri Toraman. 2020.
\newblock \href {https://dl.acm.org/doi/abs/10.1145/3397271.3401318} {Kloos: Kl
  divergence-based out-of-scope intent detection in human-to-machine
  conversations}.
\newblock In \emph{Proceedings of the 43rd International ACM SIGIR Conference
  on Research and Development in Information Retrieval}, pages 2105--2108.

\bibitem[{Yu et~al.(2019{\natexlab{a}})Yu, Zhang, Er, Li, Xue, Pang, Lin, Tan,
  Shi, Li, Jiang, Yasunaga, Shim, Chen, Fabbri, Li, Chen, Zhang, Dixit, Zhang,
  Xiong, Socher, Lasecki, and Radev}]{yu2019cosql}
Tao Yu, Rui Zhang, Heyang Er, Suyi Li, Eric Xue, Bo~Pang, Xi~Victoria Lin,
  Yi~Chern Tan, Tianze Shi, Zihan Li, Youxuan Jiang, Michihiro Yasunaga,
  Sungrok Shim, Tao Chen, Alexander Fabbri, Zifan Li, Luyao Chen, Yuwen Zhang,
  Shreya Dixit, Vincent Zhang, Caiming Xiong, Richard Socher, Walter Lasecki,
  and Dragomir Radev. 2019{\natexlab{a}}.
\newblock \href {https://www.aclweb.org/anthology/D19-1204} {{C}o{SQL}: A
  conversational text-to-{SQL} challenge towards cross-domain natural language
  interfaces to databases}.
\newblock In \emph{Proceedings of the 2019 Conference on Empirical Methods in
  Natural Language Processing and the 9th International Joint Conference on
  Natural Language Processing (EMNLP-IJCNLP)}, pages 1962--1979.

\bibitem[{Yu et~al.(2018)Yu, Zhang, Yang, Yasunaga, Wang, Li, Ma, Li, Yao,
  Roman, Zhang, and Radev}]{yu2018spider}
Tao Yu, Rui Zhang, Kai Yang, Michihiro Yasunaga, Dongxu Wang, Zifan Li, James
  Ma, Irene Li, Qingning Yao, Shanelle Roman, Zilin Zhang, and Dragomir Radev.
  2018.
\newblock \href {https://www.aclweb.org/anthology/D18-1425} {{S}pider: A
  large-scale human-labeled dataset for complex and cross-domain semantic
  parsing and text-to-{SQL} task}.
\newblock In \emph{Proceedings of the 2018 Conference on Empirical Methods in
  Natural Language Processing}.

\bibitem[{Yu et~al.(2019{\natexlab{b}})Yu, Zhang, Yasunaga, Tan, Lin, Li, Er,
  Li, Pang, Chen, Ji, Dixit, Proctor, Shim, Kraft, Zhang, Xiong, Socher, and
  Radev}]{yu2019sparc}
Tao Yu, Rui Zhang, Michihiro Yasunaga, Yi~Chern Tan, Xi~Victoria Lin, Suyi Li,
  Heyang Er, Irene Li, Bo~Pang, Tao Chen, Emily Ji, Shreya Dixit, David
  Proctor, Sungrok Shim, Jonathan Kraft, Vincent Zhang, Caiming Xiong, Richard
  Socher, and Dragomir Radev. 2019{\natexlab{b}}.
\newblock \href {https://www.aclweb.org/anthology/P19-1443} {{SP}ar{C}:
  Cross-domain semantic parsing in context}.
\newblock In \emph{Proceedings of the 57th Annual Meeting of the Association
  for Computational Linguistics}, pages 4511--4523.

\bibitem[{Zheng et~al.(2020)Zheng, Chen, and Huang}]{zheng2020out}
Yinhe Zheng, Guanyi Chen, and Minlie Huang. 2020.
\newblock \href {https://www.aclweb.org/anthology/D18-1077/} {Out-of-domain
  detection for natural language understanding in dialog systems}.
\newblock \emph{IEEE/ACM Transactions on Audio, Speech, and Language
  Processing}, pages 1198--1209.

\bibitem[{Zhong et~al.(2017)Zhong, Xiong, and Socher}]{zhong2017seq}
Victor Zhong, Caiming Xiong, and Richard Socher. 2017.
\newblock \href {https://arxiv.org/abs/1709.00103} {Seq2sql: Generating
  structured queries from natural language using reinforcement learning}.
\newblock \emph{arXiv preprint arXiv:1709.00103}.

\end{thebibliography}
